\documentclass[conference]{IEEEtran}

\usepackage{cite}
\usepackage{amsmath,amssymb,amsfonts}
\usepackage{algorithmic}
\usepackage{algorithm}
\usepackage{graphicx}
\usepackage{textcomp}
\usepackage{xcolor}
\usepackage{booktabs}
\usepackage{multirow}
\usepackage{url}
\usepackage{balance}
\usepackage{subcaption}
\usepackage{pifont}

\usepackage{enumitem}
\setlist{nosep,leftmargin=*}

\begin{document}

\title{OOM-Free Alpamayo via CPU-GPU Memory Swapping for Vision-Language-Action Models}
\author{
\IEEEauthorblockN{Seungwoo Roh, Huiyeong Kim, and Jong-Chan Kim}
\IEEEauthorblockA{
Graduate School of Automobile and Mobility, Kookmin University, Korea\\
Correspondence: jongchank@kookmin.ac.kr
}
}

\maketitle

\begin{abstract}
End-to-end Vision-Language-Action (VLA) models for autonomous driving unify perception, reasoning, and control in a single neural network, achieving strong driving performance but requiring 20--60\,GB of GPU memory---far exceeding the 12--16\,GB available on commodity GPUs.
We present a framework, which enables memory-efficient VLA inference on VRAM-constrained GPUs through system-level optimization alone, without model modification.
Our work proceeds in three stages: (1)~Sequential Demand Layering reduces VRAM usage from model-level to layer-level granularity; (2)~Pipelined Demand Layering hides parameter transfer time within layer execution time via transfer--compute overlap; and (3)~a GPU-Resident Layer Decision Policy, informed by per-module residency benefit analysis, eliminates the residual transfer overhead that pipelining cannot hide.
We further propose a performance prediction model that determines the optimal configuration---both the number and placement of resident layers---from a single profiling run with less than 1.3\% prediction error across all configurations.
Applied to NVIDIA's Alpamayo-R1-10B (21.52\,GB) on an RTX~5070\,Ti (16\,GB), our work achieves up to 3.55$\times$ speedup over Accelerate offloading while maintaining full BF16 precision.
\end{abstract}

\section{Introduction}
\label{sec:intro}

End-to-end autonomous driving increasingly adopts Vision-Language-Action (VLA) models that unify perception, reasoning, and control within a single neural network~\cite{survey_e2e_ad2024}.
Models such as Alpamayo~\cite{alpamayo2025}, RT-2~\cite{rt2_2023}, and DriveVLM~\cite{drivevlm2024} demonstrate strong driving performance, but their GPU memory requirements exceed the VRAM of commodity GPUs.
This memory gap creates a practical barrier that makes it difficult to iterate on VLA model integration, testing, and deployment without access to high-end hardware.

NVIDIA's Alpamayo-R1-10B, released in October 2025, requires 21.52\,GB of VRAM, which substantially exceeds the 12--16\,GB capacity of widely available consumer GPUs.
Consequently, inference on such GPUs is impossible on native Linux due to out-of-memory (OOM) errors. Inference on such hardware is therefore feasible only when CPU memory is jointly used with GPU memory. However, existing mechanisms for such joint use introduce prohibitive latency overhead.

One can avoid these issues by using flagship GPUs with ample VRAM such as the RTX~3090 or RTX~5090, or server-grade GPUs such as the H100, A100, or A6000.
However, these GPUs are expensive and demand consistently outstrips supply, making procurement difficult even with sufficient budget.
In this paper, we present a systematic solution, which enables optimal-performance Alpamayo inference on affordable, readily available consumer GPUs.

Model compression techniques such as quantization~\cite{gptq2023,awq2024} and pruning can reduce model size to address this problem, but these methods fundamentally incur accuracy loss.
Therefore, we target memory reduction and latency improvement through system-level optimization alone, without modifying the given model.

Our framework computes the optimal inference time achievable on a given hardware configuration through static profiling of the target model, and proceeds with optimization aimed at approaching this bound as closely as possible.
First, rather than the conventional \emph{preloading} approach that loads all layer parameters into GPU VRAM in advance, we employ \emph{Demand Layering}~\cite{demandlayering2022}, which loads parameters for each layer immediately before inference, thereby reducing VRAM usage.
Furthermore, leveraging the structural advantage that the GPU's Execution Engine and Copy Engine can simultaneously perform parameter transfer and layer inference computation, we employ \emph{pipelined Demand Layering} with a double buffer to maximally hide layer transfer time within layer execution time.
However, certain parts of the model cannot achieve full hiding.
For such modules, keeping layers GPU-resident trades VRAM cost for reduced inference latency.
We propose a policy for deciding which layers to keep resident, theoretically present a methodology for selecting resident layers, and experimentally validate this across multiple hardware platforms and models.

Our contributions are as follows:
\begin{itemize}
\item By analyzing the DMA and compute characteristics of the model's components, we derive closed-form residency benefit expressions and propose an interleaved placement policy that minimizes end-to-end inference time.
\item We present a performance prediction model that estimates inference time for arbitrary residency configurations within 1.3\% error from profiling data alone.
\item We demonstrate 3.55$\times$ speedup over baseline (Accelerate offloading) on the RTX~5070\,Ti (16\,GB) with full BF16 precision, and validate the analysis framework across multiple GPU platforms.
\end{itemize}

\section{Background}
\label{sec:background}

\subsection{Alpamayo-R1-10B Model Architecture}

Alpamayo-R1-10B~\cite{alpamayo2025} processes camera images through a 5-stage inference pipeline:

\begin{enumerate}
\item \textbf{ViT Encoder}: Qwen3-VL Vision Transformer (27 layers, $1152$ dimensions).
\item \textbf{Patch Merger}: Projects $1152{\times}4 \to 4096$ dimensions.
\item \textbf{VLM Prefill}: 36-layer Qwen3-VL-8B backbone with Grouped-Query Attention (GQA) (32Q/8KV).
\item \textbf{VLM Decode}: Autoregressive token generation---accounts for \textbf{78.5\%} of total inference time.
\item \textbf{Diffusion-based Action Generation}: Flow matching with 10 Euler steps produces 64 waypoints. The Expert Decoder (36 layers, $d{=}2048$) serves as the denoiser, called once per Euler step to predict the velocity field.
\end{enumerate}

\subsection{VRAM Decomposition: Parameter Dominance}

Parameters constitute 93\% of total memory requirements (Table~\ref{tab:vram_breakdown}).
This implies that parameter VRAM reduction is the key target.
Demand Layering~\cite{demandlayering2022} achieves memory savings by targeting precisely these parameters through layer-wise offloading.
Since VLA-based models, including Alpamayo, are similarly dominated by parameter memory, this technique is well suited for application.
Non-parameter memory accounts for the remaining 7\% and comprises the KV cache, activations, and framework buffers.
These non-parameter components are not the target of this work.

\begin{table}[t]
\centering
\caption{VRAM breakdown of Alpamayo-R1-10B (BF16).
Parameters account for 93\% of total VRAM.}
\label{tab:vram_breakdown}
\begin{tabular}{lrrr}
\toprule
\textbf{Component} & \textbf{Params} & \textbf{VRAM} & \textbf{Share} \\
\midrule
\textbf{Parameters} & \textbf{$\sim$11.08B} & \textbf{$\sim$22.16\,GB} & \textbf{93\%} \\[2pt]
\hspace{1em}Vision Encoder    & 576.4M  & 1.15\,GB  & 5.2\%  \\
\hspace{1em}VLM Backbone      & 7,580M  & 15.17\,GB & 68.5\% \\
\hspace{1em}VLM LM Head       & 637.7M  & 1.28\,GB  & 5.8\%  \\
\hspace{1em}Expert Decoder    & 2,280M  & 4.56\,GB  & 20.6\% \\[4pt]
Non-parameters                & ---     & $\sim$1.5\,GB  & 7\%  \\[2pt]
\midrule
\textbf{Total}                & ---     & \textbf{$\sim$23.7\,GB} & \textbf{100\%} \\
\bottomrule
\end{tabular}

\vspace{2pt}
\raggedright\footnotesize
The 21.52\,GB cited elsewhere is \texttt{max\_memory\_allocated} (RTX~3090); the 23.7\,GB total includes alignment and framework buffers.
\end{table}

\subsection{PCIe DMA and CUDA Streams}

Modern NVIDIA GPUs contain independent Copy Engines and Execution Engines~\cite{cuda_programming_guide}.
The CE handles PCIe DMA transfers while the EE executes CUDA kernels.
Transfer--compute overlap is achievable by submitting workloads on separate CUDA streams.

\begin{figure}[t]
\centering


\begin{subfigure}[b]{\columnwidth}
\centering
\includegraphics[width=\columnwidth]{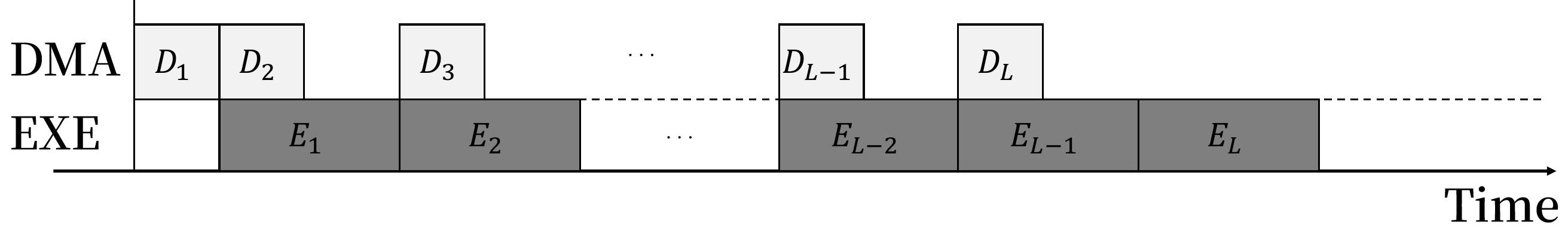}
\caption{$C_\text{DMA} < C_\text{EXE}$: DMA is fully hidden within the execution window.}
\label{fig:overlap_exe_intensive}
\end{subfigure}

\bigskip
\bigskip

\begin{subfigure}[b]{\columnwidth}
\centering
\includegraphics[width=\columnwidth]{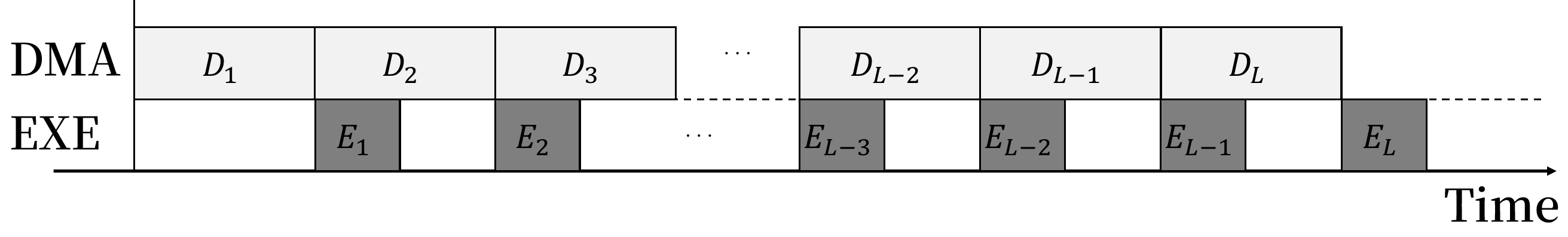}
\caption{$C_\text{DMA} > C_\text{EXE}$: DMA becomes the bottleneck; pipelining cannot fully hide transfers.}
\label{fig:overlap_dma_intensive}
\end{subfigure}
\caption{Inference timelines under two regimes. (a)~EXE-intensive pipelining where transfers are fully hidden. (b)~DMA-intensive pipelining where transfer overhead is unavoidable.}
\label{fig:overlap}
\end{figure}

Fig.~\ref{fig:overlap} illustrates transfer--compute overlap.
If the PCIe DMA parameter transfer time for a given layer is shorter than the layer execution time, the parameter transfer for layer~$i$ can complete before the computation of layer~$i{-}1$ finishes.
Let $C^{i}_{\text{DMA}}$ and $C^{i}_{\text{EXE}}$ denote the DMA transfer time and execution time of layer $i$, respectively.
As shown in Fig.~\ref{fig:overlap}(a), the total inference time for all $L$ layers is:
\begin{equation}
C^{1}_{\text{DMA}} + \sum_{i=1}^{L} C^{i}_{\text{EXE}}.
\label{eq:overlap_dma_lt_exe}
\end{equation}

If the parameter transfer time exceeds the layer execution time, the situation corresponds to Fig.~\ref{fig:overlap}(b).
In this case, parameter transfers cannot be fully hidden, and the total inference time becomes:
  \begin{equation}                                                                                                 
  \sum_{i=1}^{L} C^{i}_{\text{DMA}} + C^{L}_{\text{EXE}}.
  \label{eq:overlap_dma_gt_exe}
  \end{equation}   

When GPU VRAM is sufficient for all parameters, the model operates in a preloading mode with no runtime parameter loading required, yielding an inference time of $\sum_{i} C^{i}_{\text{EXE}}$.
Relative to this baseline, the overhead introduced by Demand Layering depends on each layer's characteristics.


To simplify the analysis, we first assume that all layers within a module share the same characteristic---either $C_{\text{DMA}} < C_{\text{EXE}}$ or $C_{\text{DMA}} > C_{\text{EXE}}$ holds uniformly across all layers.

When $\text{DMA} < \text{EXE}$, parameter transfers can be fully hidden, incurring only the overhead of the initial layer's transfer time $C^{1}_{\text{DMA}}$.
In contrast, when $\text{DMA} > \text{EXE}$, parameter loading cannot be fully hidden, and each layer incurs an overhead of $C_{\text{DMA}} - C_{\text{EXE}}$.
In this case, DMA becomes the bottleneck, pipelining cannot completely hide the transfer, and the only solution is GPU residency to eliminate the transfer entirely.

In practice, a model may contain layers of both types. The total overhead is then a per-layer combination of the two expressions above.

\section{Problem Analysis}
\label{sec:problem}

We identify three challenges in reducing Alpamayo's inference latency: (i)~theoretical lower bound, (ii)~overhead of existing swapping methods, and (iii)~DMA hiding limitation.

\subsection{Theoretical Lower Bound}

The RTX~5090 has 21,760 CUDA cores and the RTX~5070\,Ti has 8,960 CUDA cores.
Therefore, even if the RTX~5070\,Ti had 24\,GB or more of VRAM, its inference time would still be longer due to the GPU's compute capability.
We define the \emph{theoretical lower bound} as the time required to perform Alpamayo inference with infinite VRAM.
No offloading strategy can achieve inference latency below this bound.
GPUs with ample VRAM can preload all parameters for inference, but for GPUs like the RTX~5070\,Ti whose VRAM is smaller than the model parameter size, preloading is impossible.
We therefore estimate the theoretical lower bound as the simple sum of per-layer execution times:
\begin{equation}
C^{\text{opt}} = \sum_{i=1}^{L} C^{i}_{\text{EXE}}.
\label{eq:theoretical_lower_bound}
\end{equation}
This estimate excludes non-layer overhead (KV cache operations, Python dispatch, CUDA launch latency), which is empirically negligible: on the RTX~3090, the measured preloading time of 3.12\,s differs from the estimated $C^{\text{opt}} = 3.13$\,s by only 0.3\%.

\begin{table}[t]
\centering
\caption{Measured inference time and theoretical lower bound estimated from per-layer execution times.}
\label{tab:theoretical}
\begin{tabular}{lrr}
\toprule
\textbf{GPU} & $C^{\text{opt}}$ & \textbf{Measured Time} \\
\midrule
RTX 5070 Ti & 2.86\,s & -- \\
RTX 3090    & 3.13\,s & 3.12\,s \\
\bottomrule
\end{tabular}
\end{table}

\subsection{Swapping Overhead in Existing Methods}

For GPUs such as the RTX~5070\,Ti that cannot preload all model parameters, an efficient parameter swapping strategy that minimizes the additional overhead is essential.

When inferring a model that exceeds GPU VRAM without explicit swapping, the behavior differs by operating system.
The WDDM driver used in Windows and WSL2 environments transparently migrates data between VRAM and system RAM in 4\,KB--2\,MB page granularity through GPU page faults.
As a result, Alpamayo-R1-10B inference takes 273.79\,s on an RTX~3080\,Ti (12\,GB) and 69.6\,s on an RTX~5070\,Ti (16\,GB).
The KMD driver on native Linux raises an OOM error instead of performing transparent swapping, making an explicit offloading framework mandatory.

The Hugging Face Accelerate library~\cite{accelerate2022} provides the blind
swapping method.
This mechanism statically partitions each parameter tensor across GPU and CPU at model load time, then synchronously transfers tensors to the GPU via forward hooks during inference, deleting them afterward.
This approach has three fundamental inefficiencies:

\begin{itemize}
\item \textbf{Per-parameter transfer}: In a transformer layer, 14 weight tensors are transferred via individual \texttt{.to("cuda")} calls, resulting in 14 malloc-H2D operations per layer.
\item \textbf{Synchronous execution}: The CPU thread blocks until each parameter arrives. The H2D transfer of layer~$i{+}1$ cannot begin until the forward pass of layer~$i$ completes.
\item \textbf{Global synchronization}: Module deletion triggers \texttt{empty\_cache()} or block reclamation, causing global CUDA stream synchronization.
\end{itemize}

\begin{figure}[t]                                                                                                
\centering
\includegraphics[width=\columnwidth]{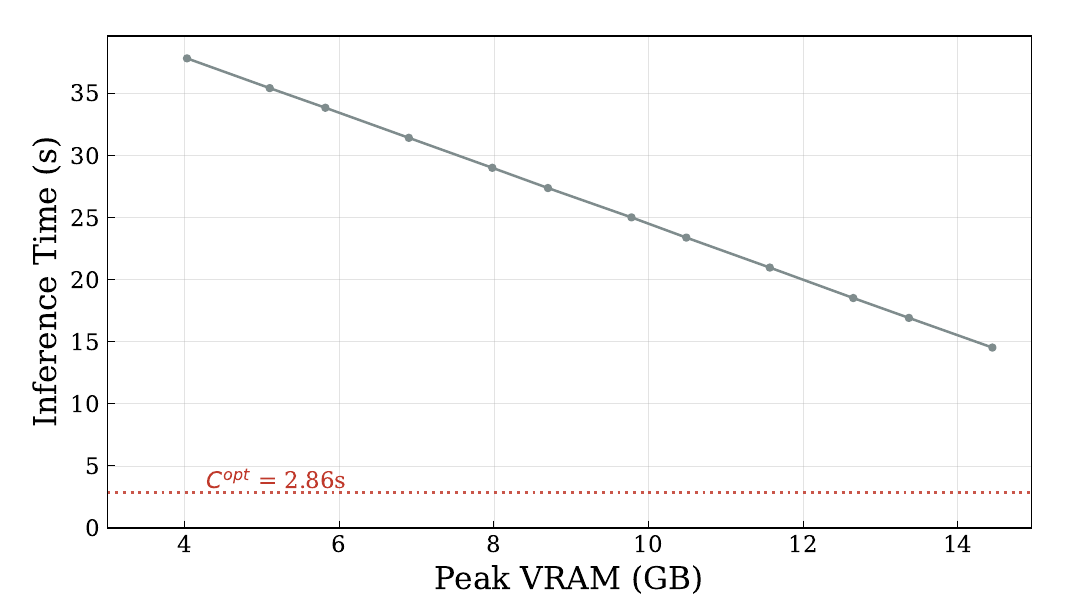}                                                  
\caption{VRAM--latency tradeoff of Hugging Face Accelerate offloading under varying VRAM limits on RTX~5070\,Ti.}     
\label{fig:devmap_tradeoff}                                                                                      
\end{figure}

On the RTX~5070\,Ti, Hugging Face Accelerate offloading incurs an inference latency of 14.52\,s, which substantially exceeds the theoretical lower bound.
Moreover, this gap widens drastically as the available VRAM is further constrained (Fig.~\ref{fig:devmap_tradeoff}).

\subsection{DMA Hiding Limitation}

The primary source of swapping overhead is the transfer time for parameter data.
Since transfers occur from CPU memory to GPU VRAM over PCIe, this overhead is determined by PCIe transfer performance---a hardware limitation.
Applying pipelining with Demand Layering can hide this transfer time when a layer's execution time exceeds its parameter transfer time.
However, in Alpamayo, some modules have parameter transfer times that are much larger than their layer execution times (Table~\ref{tab:per_layer_profile}).
We define the DMA-to-EXE ratio $r = C_{\text{DMA}} / C_{\text{EXE}}$ for each module (Table~\ref{tab:per_layer_profile}). Modules with $r < 1$ are EXE-intensive (DMA fully hideable), while $r > 1$ indicates DMA-intensive behavior where pipelining alone cannot eliminate the transfer overhead.
In such cases, hiding the transfer time is structurally impossible, and selecting GPU-resident layers from the remaining VRAM to eliminate transfer time entirely becomes essential.

  \begin{table}[t]                                                                                                 
  \centering                                                                                                       
  \caption{Per-layer DMA and execution cost (RTX~5070\,Ti, Sequential Demand Layering, 30 iterations, pinned memory, PCIe Gen5).}                                                                                                          
  \label{tab:per_layer_profile}
  \begin{tabular}{lccc}
  \toprule
  Module & $C_{\text{DMA}}$ (ms) & $C_{\text{EXE}}$ (ms) & $r$ \\
  \midrule
  Vision       & 0.9  & 8.3  & 0.11 \\
  VLM Prefill  & 10.9 & 16.6 & 0.66 \\
  VLM Decode   & 10.9 & 0.9  & 12.1 \\
  Diffusion    & 3.6  & 1.0  & 3.5  \\
  \bottomrule                                                                                                           
  \end{tabular}                                                   
  \end{table} 

\section{System Architecture}
\label{sec:framework}

\subsection{Design Overview}

Our work aims to minimize inference latency on VRAM-constrained commodity GPUs, approaching the theoretical lower bound as closely as possible.
The theoretical lower bound can be measured directly on GPUs with sufficient VRAM by running Alpamayo inference with all parameters preloaded.
For GPUs such as the RTX~5070\,Ti where preloading is infeasible due to VRAM constraints, the bound is estimated as the sum of per-layer execution times (Eq.~\eqref{eq:theoretical_lower_bound}), since non-layer overhead is negligible.

To approach this theoretical lower bound (Table~\ref{tab:theoretical}), we proceed in three optimization stages:
\begin{enumerate}
\item \textbf{Sequential Demand Layering}: Reduces VRAM usage from model-level to layer-level by loading parameters on demand.
\item \textbf{Pipelined Demand Layering}: Hides parameter transfer time within layer execution time using transfer--compute overlap.
\item \textbf{GPU-Resident Layer Selection}: Allocates the freed VRAM to keep selected layers GPU-resident, eliminating the transfer time that pipelining could not fully hide.
\end{enumerate}

\subsection{Sequential Demand Layering}

Demand Layering~\cite{demandlayering2022} loads each layer's parameters immediately before its inference, as opposed to the conventional preloading approach that uploads all layer parameters to VRAM before the first inference cycle.
While Demand Layering originally targeted CNNs with their sequential execution characteristics, Alpamayo is similarly well suited for this technique because parameters dominate its total VRAM consumption.

The most significant contribution of Demand Layering~\cite{demandlayering2022} was reducing parameter memory from model-level to layer-level granularity.
Our framework likewise introduces per-layer on-demand parameter loading, reducing VRAM from 14.45\,GB to 3.99\,GB---a 72\% reduction.
By addressing the first inefficiency of Hugging Face Accelerate offloading---per-variable sequential allocation and loading within each layer---we consolidate parameter transfers at layer granularity.
Despite the structural limitation that parameter transfer and layer execution remain serialized in this sequential variant, execution time is reduced from 14.52\,s to 12.72\,s.

\subsection{Pipelined Demand Layering}

GPUs possess both an Execution Engine and a Copy Engine, enabling concurrent layer inference (EXE) and parameter copying from CPU memory to GPU VRAM (DMA).
We exploit this structural characteristic by applying pipelining to Demand Layering.

The purpose of pipelining is to hide parameter loading within layer execution time.
We classify layers based on whether the parameter transfer time or the execution time dominates:
\begin{itemize}
\item \textbf{EXE-intensive modules}: $C_{\text{DMA}} < C_{\text{EXE}}$. DMA is fully hidden within EXE; overhead relative to preloading is a single $C_{\text{DMA}}$ for the first layer.
\item \textbf{DMA-intensive modules}: $C_{\text{DMA}} > C_{\text{EXE}}$. DMA cannot be fully hidden; each layer incurs overhead of $C_{\text{DMA}} - C_{\text{EXE}}$, yielding total overhead of $L \cdot (C_{\text{DMA}} - C_{\text{EXE}})$.
\end{itemize}

\begin{figure}[t]
\centering
\includegraphics[width=\columnwidth]{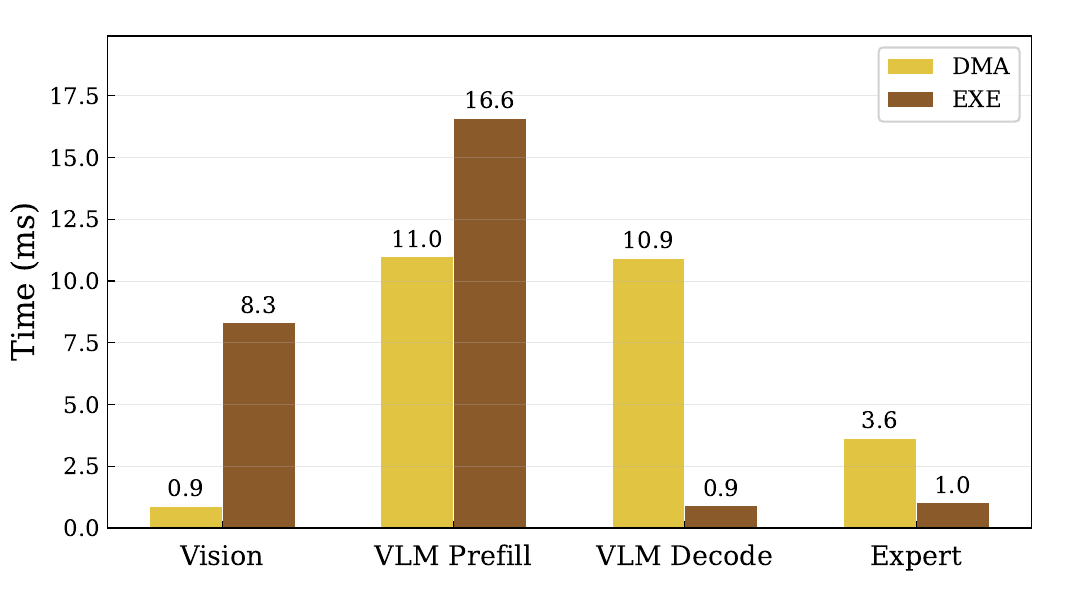}
\caption{Average per-layer DMA and execution times for each Alpamayo module (RTX~5070\,Ti). ViT and VLM Prefill are EXE-intensive ($C_\text{DMA} < C_\text{EXE}$); VLM Decode and Expert are DMA-intensive ($C_\text{DMA} > C_\text{EXE}$).}
\label{fig:module_dma_exe_profile}
\end{figure}

The CNNs targeted by Demand Layering~\cite{demandlayering2022} have EXE-intensive module behavior, enabling substantial memory savings at the cost of only a single $C_{\text{DMA}}$ overhead.
However, Alpamayo's modules exhibit heterogeneous characteristics.
We measured per-layer parameter transfer times and execution times for each Alpamayo module (Fig.~\ref{fig:module_dma_exe_profile}).
The results show that the ViT module and the VLM prefill phase are EXE-intensive, while the VLM decode phase and the Diffusion module are DMA-intensive.

DMA for the ViT module and the VLM prefill phase can be fully hidden by adjusting pipelining, and inference time can approach the theoretical lower bound. However, DMA for the VLM decode phase and the Diffusion module cannot be fully hidden. Thus, additional measures are required to reduce latency.

To implement pipelined Demand Layering, we introduce the \emph{Double Flat Buffer} (DFB), a GPU memory management scheme that enables concurrent DMA transfer and layer execution without per-layer memory allocation overhead. The DFB comprises three design elements: (i) Buffer structure, (ii) Slot assignment and synchronization, and (iii) Parameter pinning.


\textbf{(i) Buffer structure.} The DFB allocates two contiguous GPU buffers (slots 0 and 1), each sized to hold the largest layer's parameters across all modules.
Rather than allocating and freeing GPU memory for each layer via \texttt{torch.empty()} and \texttt{empty\_cache()}, all layers share the same two pre-allocated buffers in an alternating fashion.
This eliminates the \texttt{empty\_cache()} synchronization overhead that dominates Hugging Face Accelerate offloading.

All modules (ViT, VLM, Expert) share a single unified DFB instance, with each buffer slot sized to the largest layer across all modules.
The total buffer memory is $2 \times M_{\max}$, where $M_{\max}$ is the largest layer size among all modules.


\textbf{(ii) Slot assignment and synchronization.} Layers are assigned to slots via $s = i \bmod 2$, where $i$ is the layer index.
Two CUDA events govern the pipeline:

\begin{itemize}
\item \textbf{DMA-done event}: Recorded on the prefetch stream after H2D copy completes. The default (compute) stream waits on this event before executing the layer.
\item \textbf{Compute-done event}: Recorded on the default stream after layer execution completes. The prefetch stream waits on this event before reusing the slot for the next layer's DMA.
\end{itemize}

This two-event protocol ensures that (1)~a layer never executes before its parameters arrive, and (2)~a buffer slot is never overwritten while its data is still in use.
The prefetch stream submits the DMA for layer~$i{+}1$ while the default stream executes layer~$i$, achieving the transfer--compute overlap illustrated in Fig.~\ref{fig:overlap}(a)--(b).


\textbf{(iii) Parameter pinning.} All CPU-side layer parameters are registered as CUDA pinned (page-locked) memory via \texttt{pin\_memory()}.
Pinned memory enables the GPU's Copy Engine to perform DMA transfers via \texttt{non\_blocking=True} without CPU involvement, which is essential for achieving full CE/EE overlap.
On the RTX~5070\,Ti (PCIe Gen5), pinned H2D throughput reaches 33.4\,GB/s.

\subsection{GPU-Resident Layer Decision Policy}
\label{subsec:resident_policy}

Sequential Demand Layering reduces VRAM usage to layer-level granularity, and pipelined Demand Layering hides transfer time for EXE-intensive modules.
However, DMA-intensive modules cannot achieve full transfer hiding.
The \emph{GPU-resident layer decision policy} addresses this limitation: keeping a specific layer GPU-resident eliminates its DMA time at the cost of permanently occupying VRAM equal to that layer's parameter size.

The efficiency of GPU residency varies across modules.
We define the \emph{residency benefit} of a layer as the inference time saved by keeping it resident, divided by its memory footprint.
For modules whose layers are invoked repeatedly---e.g., VLM decode layers are invoked once per output token during autoregressive generation, and diffusion layers are invoked once per diffusion step---the time savings are multiplied by the repetition count:

\begin{equation}
B_{\ell} = \frac{R_{\ell} \cdot \Delta C_{\ell}}{M_{\ell}}.
\label{eq:residency_benefit}
\end{equation}
where $B_{\ell}$ is the residency benefit of layer~$\ell$, $R_{\ell}$ is its repetition count, $\Delta C_{\ell} = C^{\ell}_{\text{DMA}} - \min(C^{\ell}_{\text{DMA}}, C^{\ell}_{\text{EXE}})$ is the per-invocation time saved, and $M_{\ell}$ is its memory footprint.

Before computing each module's residency benefit, we measured per-layer DMA and execution times and confirmed that layers within the same module are functionally identical in code structure and exhibit identical measured timings (Fig.~\ref{fig:module_dma_exe_profile}).
We now derive the residency benefit based on each module's characteristics, and then apply it to Alpamayo's specific modules.

\begin{figure}[t]                                                                                                
\centering                                                                                                       
\includegraphics[width=\columnwidth]{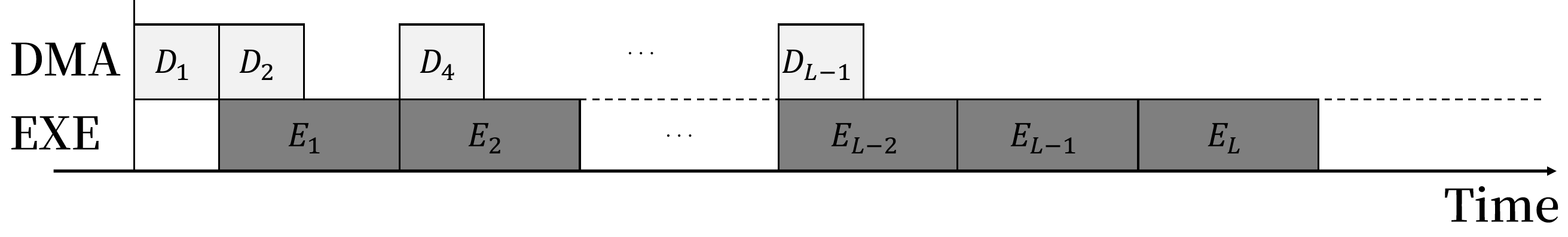}                                               
\caption{Pipelined EXE-intensive module with layer~3 resident. Removing $D_2$ leaves an idle DMA slot, but total 
time is unchanged since EXE dominates---confirming zero benefit for middle-layer residency.}                     
\label{fig:exe_intensive_p2}                                                                                     
\end{figure}

\textbf{EXE-Intensive Modules.} Consider a module with $L$ layers where $C_{\text{DMA}} < C_{\text{EXE}}$, invoked $R$ times per inference.
Pipelining fully hides all DMA transfers except the initial one per invocation, yielding:
\begin{equation}
C_{\text{EXE\text{-}int}} = R \cdot (C_{\text{DMA}} + L \cdot C_{\text{EXE}}).
\label{eq:exe_intensive_time}
\end{equation}
Fig.~\ref{fig:exe_intensive_p2} illustrates this pipeline: since $C_\text{EXE} > C_\text{DMA}$, removing a middle layer's DMA (e.g., $D_2$) leaves an idle slot without reducing total time.
Keeping the first layer resident removes the initial DMA across all $R$ invocations in Eq.~\eqref{eq:exe_intensive_time}.
Keeping any middle or last layer resident provides no benefit, as its DMA is already hidden within the preceding layer's EXE.
Table~\ref{tab:exe_intensive_benefit} summarizes the residency benefit by position.

\begin{table}[t]
\centering
\caption{EXE-intensive module ($R$ repetitions): residency benefit by position.}
\label{tab:exe_intensive_benefit}
\begin{tabular}{lll}
\toprule
\textbf{Position} & $\Delta C$ & \textbf{Benefit} $B$ \\
\midrule
First & $R \cdot C_{\text{DMA}}$ & $R \cdot C_{\text{DMA}} / M$ \\
Middle & $0$ & $0$ \\
Last & $0$ & $0$ \\
\bottomrule
\end{tabular}
\end{table}

\begin{figure}[t]                                                                                                
\centering                                                      
\includegraphics[width=\columnwidth]{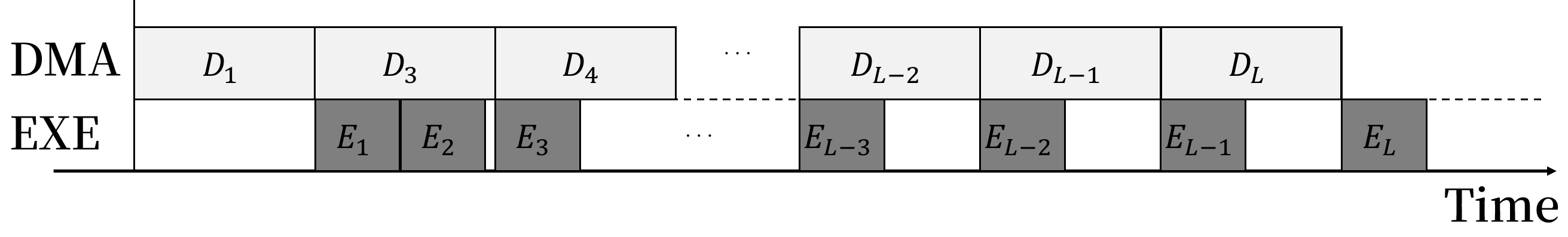}
\caption{Pipelined DMA-intensive module with layer 2 resident. DMA dominates the timeline ($C_\text{DMA} > C_\text{EXE}$), and each
layer incurs an irreducible overhead of $C_\text{DMA} - C_\text{EXE}$ that pipelining cannot hide.}              
\label{fig:dma_intensive_p1}
\end{figure}

\textbf{DMA-Intensive Modules.} Consider a module with $L$ layers where $C_{\text{DMA}} > C_{\text{EXE}}$, invoked $R$ times per inference (e.g., once per output token or per diffusion step).
Under pipelining, all $L$ layers' DMA transfers form the bottleneck, and only the last layer's execution cannot be overlapped (the ``trailing EXE'').
The total execution time is therefore:
\begin{equation}
C_{\text{DMA\text{-}int}} = R \cdot (L \cdot C_{\text{DMA}} + C_{\text{EXE}}),
\label{eq:dma_intensive_time}
\end{equation}
where the $C_{\text{EXE}}$ term is the single trailing execution of the last layer per invocation (Fig.~\ref{fig:dma_intensive_p1}).
This applies to both VLM Decode ($R{=}N$ tokens, $L{=}36$) and the Diffusion Expert ($R{=}10$ steps, $L{=}36$).

Keeping a layer resident eliminates its $C_{\text{DMA}}$ across all $R$ invocations in Eq.~\eqref{eq:dma_intensive_time}.
For the last layer, the benefit is reduced because DMA of the next iteration's first layer can partially overlap with its EXE.
The residency benefit by position is summarized in Table~\ref{tab:dma_intensive_benefit}.

\begin{table}[t]
\centering
\caption{DMA-intensive module ($R$ repetitions): residency benefit by position.}
\label{tab:dma_intensive_benefit}
\begin{tabular}{lll}
\toprule
\textbf{Position} & $\Delta C$ & \textbf{Benefit} $B$ \\
\midrule
First / Middle & $R \cdot C_{\text{DMA}}$ & $R \cdot C_{\text{DMA}} / M$ \\
Last & $R \cdot (C_{\text{DMA}} {-} C_{\text{EXE}})$ & $R \cdot (C_{\text{DMA}} {-} C_{\text{EXE}}) / M$ \\
\bottomrule
\end{tabular}
\end{table}

When consecutive layers are kept resident, the DMA interval for the next offloaded layer can overlap with the execution of multiple resident layers.
The maximum number of consecutive resident layers whose EXE can be hidden within a single DMA is $\lfloor C_{\text{DMA}} / C_{\text{EXE}} \rfloor$; we call this the \emph{consecutive residency limit}.


\textbf{Hybrid Modules.} When the same parameters serve both an EXE-intensive phase and a DMA-intensive phase, keeping a layer resident benefits both.
Since VLM Prefill is EXE-intensive ($C_{\text{DMA}} < C_{\text{prefill}}$) and VLM Decode is DMA-intensive ($C_{\text{DMA}} > C_{\text{decode}}$), each phase uses its respective pipelined formula (Eq.~\eqref{eq:overlap_dma_lt_exe} for prefill, Eq.~\eqref{eq:overlap_dma_gt_exe} for decode): leading DMA + all EXE for prefill, and all DMA + trailing EXE for decode.
Let the DMA-intensive phase repeat $N$ times.
The full-offload execution time is:
\begin{equation}
C_{\text{hybrid}} = C_{\text{DMA}} + L \cdot C_{\text{prefill}} + N \cdot (L \cdot C_{\text{DMA}} + C_{\text{decode}}).
\label{eq:hybrid_time}
\end{equation}

The combined benefit is the sum of the EXE-intensive first-layer savings and the DMA-intensive repeated savings, as shown in Table~\ref{tab:hybrid_benefit}.

\begin{table}[t]
\centering
\caption{Hybrid module ($N$ decode repetitions): residency benefit by position.}
\label{tab:hybrid_benefit}
\begin{tabular}{lll}
\toprule
\textbf{Position} & $\Delta C$ & \textbf{Benefit} $B$ \\
\midrule
First & $(N{+}1) \cdot C_{\text{DMA}}$ & $(N{+}1) \cdot C_{\text{DMA}} / M$ \\
Middle & $N \cdot C_{\text{DMA}}$ & $N \cdot C_{\text{DMA}} / M$ \\
Last & $N \cdot (C_{\text{DMA}} {-} C_{\text{decode}})$ & $N \cdot (C_{\text{DMA}} {-} C_{\text{decode}}) / M$ \\
\bottomrule
\end{tabular}
\end{table}

The last layer's reduced benefit warrants explanation: in the full-offload pipeline, the last layer's execution serves as the ``trailing EXE'' that is already excluded from the pipelined DMA chain (the $C_{\text{decode}}$ term in Eq.~\eqref{eq:hybrid_time}).
When the last layer becomes resident and its DMA is eliminated, the \emph{penultimate} layer becomes the new pipeline tail, introducing a new trailing execution cost of $C_{\text{decode}}$ that was previously absorbed.
Consequently, the net savings is only $N \cdot (C_{\text{DMA}} - C_{\text{decode}})$ rather than the full $N \cdot C_{\text{DMA}}$.


\label{subsubsec:module_benefit}

We now apply this framework to Alpamayo's modules. We instantiate the general framework for each Alpamayo module using the profiling data from Table~\ref{tab:per_layer_profile}.
Non-layer components (Patch Embed, Patch Merger, LM-Head) have higher per-MB benefit than any layer and are kept resident in all configurations.

\begin{figure}[t]                                                                                                
\centering                                                         
\includegraphics[width=\columnwidth]{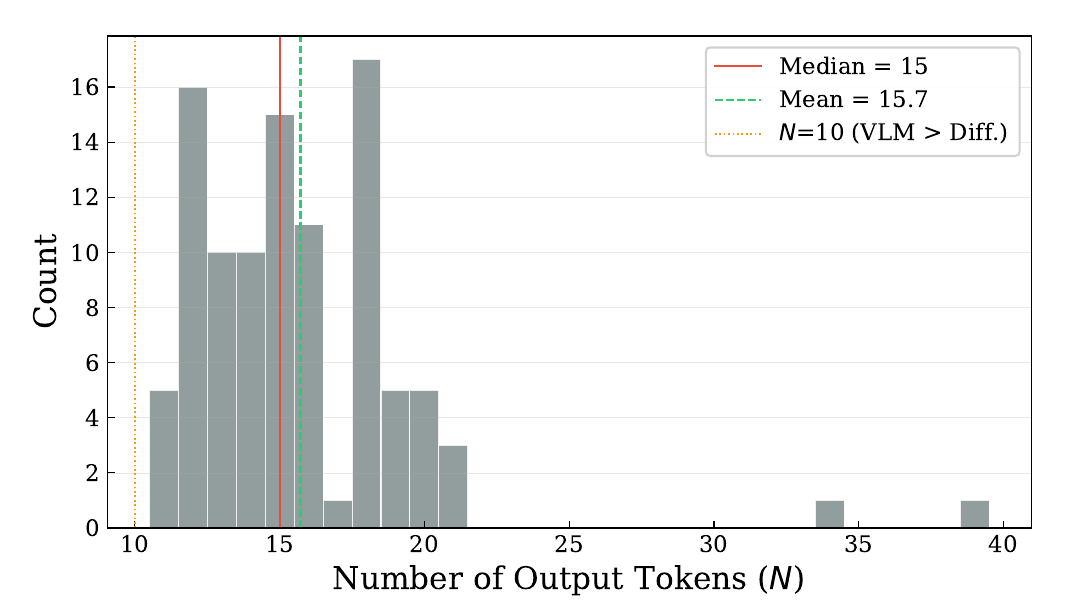}
\caption{Distribution of VLM output token counts across 100 randomly sampled driving clips. All inputs produce $N
\ge 11$ tokens ($\text{median}{=}15$, $\text{mean}{=}15.7$), well above the $N{=}10$ threshold where VLM        
residency benefit exceeds Diffusion.}
\label{fig:token_distribution}                                                                                   
\end{figure}

\begin{itemize}
\item \textbf{ViT Encoder} (27 layers, $M{=}29.1$\,MB): EXE-intensive ($r{=}0.11$, $R{=}1$).
$B_{\text{first}} = 1 \times 0.9 / 29.1 = 0.031$\,ms/MB; $B_{\text{mid}} = B_{\text{last}} = 0$.

\item \textbf{VLM Backbone} (36 layers, $M{=}368.0$\,MB): Hybrid. Prefill is EXE-intensive ($r{=}0.66$); Decode is DMA-intensive ($r{=}12.1$, $N{=}21$ tokens).
We fix $N{=}21$ throughout our analysis, matching the reference evaluation clip; this is a conservative choice, as the empirically measured median is 15 tokens (Fig.~\ref{fig:token_distribution}).
$B_{\text{first}} = 22 \times 10.9 / 368.0 = 0.651$\,ms/MB;
$B_{\text{mid}} = 21 \times 10.9 / 368.0 = 0.622$\,ms/MB;
$B_{\text{last}} = 21 \times (10.9 - 0.9) / 368.0 = 0.571$\,ms/MB.

\item \textbf{Expert Decoder} (36 layers, $M{=}120.8$\,MB): DMA-intensive ($r{=}3.5$, $R{=}10$ steps).
$B_{\text{first}} = B_{\text{mid}} = 10 \times 3.6 / 120.8 = 0.298$\,ms/MB;
$B_{\text{last}} = 10 \times (3.6 - 1.0) / 120.8 = 0.215$\,ms/MB.
\end{itemize}

Table~\ref{tab:module_benefit_summary} summarizes the computed benefits.
VLM layers dominate across all positions, confirming that optimal residency allocation should prioritize VLM layers.

\begin{table}[t]
\centering
\caption{Computed residency benefit for each Alpamayo module (RTX~5070\,Ti, $N{=}21$).}
\label{tab:module_benefit_summary}
\begin{tabular}{lcccc}
\toprule
\textbf{Module} & \textbf{Type} & $B_{\text{first}}$ & $B_{\text{mid}}$ & $B_{\text{last}}$ \\
 & & \multicolumn{3}{c}{(ms/MB)} \\
\midrule
ViT         & EXE-int  & 0.031  & 0     & 0     \\
VLM         & Hybrid   & 0.651  & 0.622 & 0.571 \\
Expert      & DMA-int  & 0.298  & 0.298 & 0.215 \\
\bottomrule
\end{tabular}
\end{table}

\begin{figure}[t]
\centering
\includegraphics[width=\columnwidth]{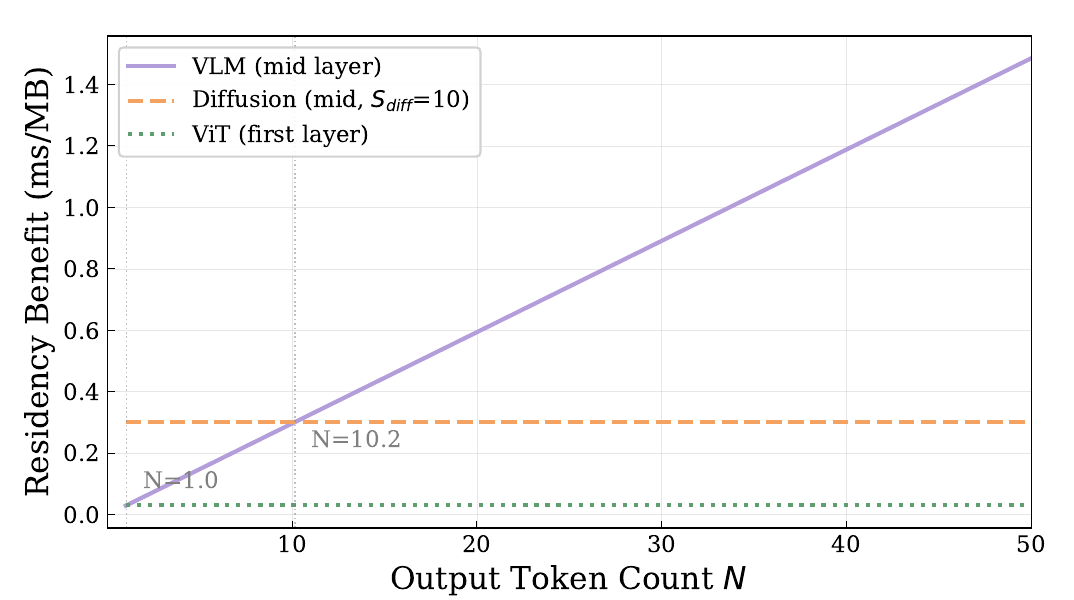}
\caption{Residency benefit per module versus output token count ($S_\text{diff}{=}10$). VLM benefit exceeds Diffusion at $N{>}10$ and dominates at Alpamayo's typical $N{=}21$.}
\label{fig:residency_benefit_vs_tokens}
\end{figure}

Fig.~\ref{fig:residency_benefit_vs_tokens} shows the residency benefit of each module's layers as a function of the output token count~$N$ with diffusion steps fixed at~10.
Computing the crossover points reveals that VLM residency benefit surpasses ViT and Diffusion benefit when the output token count exceeds certain thresholds.

Fig.~\ref{fig:token_distribution} shows the distribution of output token counts across 100 randomly sampled clips. The median is 15 tokens with P5--P95 range of 12--21, confirming that VLM residency benefit dominates in virtually all driving scenarios. The empirical distribution confirms that ViT and Diffusion layers have lower residency benefit than VLM layers in the vast majority of cases. Furthermore, on the RTX~5070\,Ti, VRAM is insufficient to keep all VLM layers resident.
We therefore conclude that ViT and Diffusion layers should always be offloaded, and optimal residency allocation should focus exclusively on VLM layers.

\begin{figure}[t]
\centering
\includegraphics[width=\columnwidth]{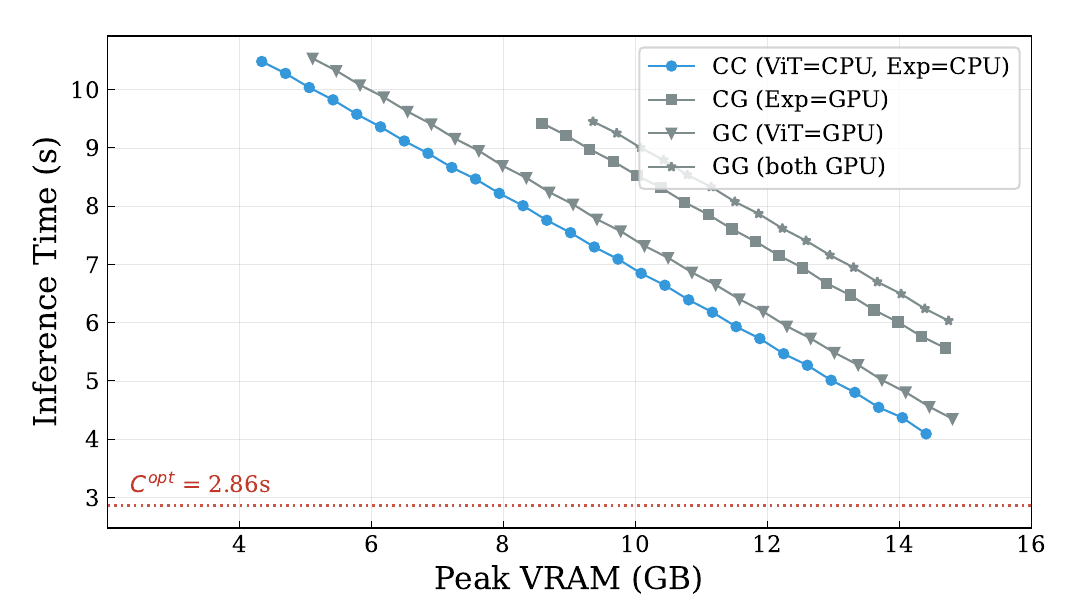}
\caption{Module residency comparison. Keeping ViT and Expert offloaded (CC) yields the lowest latency for any given VRAM budget, confirming the theoretical residency benefit analysis.}
\label{fig:resident_vlm_sweep}
\end{figure}

Fig.~\ref{fig:resident_vlm_sweep} shows inference time as a function of the number of GPU-resident VLM layers, comparing configurations where ViT and Diffusion modules are kept resident versus offloaded.
Inference time decreases linearly with the number of resident VLM layers.
Consistent with the residency benefit analysis, offloading ViT and Diffusion modules yields superior inference performance for the same VRAM budget, confirming the theoretical prediction experimentally.


Having confirmed the residency benefit experimentally, we now determine the optimal resident layer configuration.
Na\"ively, selecting which layers to keep resident from 36~VLM layers requires evaluating $2^{36}$ possible configurations.
However, since all layers within the same module are functionally identical and exhibit uniform DMA and execution times, the residency benefit depends only on the \emph{number} of resident layers, not on their specific identities.
This reduces the search space from $2^{36}$ to at most 37 configurations (0 through 36 resident layers).

The remaining question is \emph{which} $k$ layers to select when keeping $k$ layers resident.
Three observations guide this decision:

\begin{enumerate}
\item \textbf{First layer inclusion}: The first layer provides strictly higher benefit than middle layers ($(N{+}1) \cdot C_{\text{DMA}}$ vs.\ $N \cdot C_{\text{DMA}}$), so it is always included.
\item \textbf{Last layer exclusion}: The last layer provides the lowest benefit ($N \cdot (C_{\text{DMA}} - C_{\text{decode}})$), so it is always excluded.
\item \textbf{Consecutive residency limit}: As discussed in Section~\ref{subsec:resident_policy}, when more than $\lfloor C_{\text{DMA}} / C_{\text{decode}} \rfloor \approx 12$ consecutive layers are kept resident, the DMA interval of the next offloaded layer can no longer fully overlap with the execution of the resident layers, causing a marginal loss in residency benefit.
\end{enumerate}

While consecutive and interleaved placement yield similar overall performance, interleaved placement avoids the consecutive residency limit entirely.
We therefore adopt the following \emph{interleaved residency placement} policy: given $k$ layers to keep resident among layers $0$ through $34$ (first included, last excluded), the resident layers are placed at evenly spaced indices to maximize the gap between consecutive resident layers.
Formally, the $k$ resident layer indices are:
\begin{equation}
\mathcal{R}(k) = \left\{ \left\lfloor \frac{i \cdot 35}{k} \right\rfloor \ \middle|\ i = 0, 1, \ldots, k{-}1 \right\}.
\label{eq:interleaved_placement}
\end{equation}

Eq.~\eqref{eq:interleaved_placement} ensures that the inter-resident gap remains at least $\lfloor 35/k \rfloor$, which stays above the consecutive residency limit for all practical $k$ values on the RTX~5070\,Ti.
Combined with the performance prediction model, this enables our method to determine the optimal configuration---both the number and placement of resident layers---from a single profiling run without exhaustive search.

\subsection{Performance Prediction Model}
\label{subsec:prediction}

The residency benefit $B_\ell$ defined in Eq.~\eqref{eq:residency_benefit} directly yields a performance prediction model.
Since inference time decreases linearly with the number of resident VLM layers, the entire VRAM--latency trade-off curve can be predicted from profiling data and a single measurement.


\textbf{Prediction Model.} Since the first layer is always resident and the last layer is always excluded, the $k$ additional resident layers are all middle layers with uniform benefit $B_\text{VLM} = B_\text{mid}$.
Given $B_\text{VLM}$ and the memory footprint $M_\text{VLM}$ of a single VLM layer, the inference time with $k$ GPU-resident VLM layers is:
\begin{equation}
C_{\text{total}}(k) = C_{\text{total}}(0) - k \cdot B_\text{VLM} \cdot M_\text{VLM},
\label{eq:sweep_prediction}
\end{equation}
where $C_{\text{total}}(0)$ is the full-offload inference time measured on the final system.
Equivalently, for a given VRAM budget $V_\text{resident}$ allocated to resident layers:
\begin{equation*}
C_{\text{total}}(V) = C_{\text{total}}(0) - B_\text{VLM} \cdot V_\text{resident}.
\end{equation*}

The slope of the linear model in Eq.~\eqref{eq:sweep_prediction} is $-B_\text{VLM} \cdot M_\text{VLM}$, which is computed entirely from profiling data ($C_\text{DMA}$, $C_\text{EXE}$, repetition count, and layer size).
Only $C_{\text{total}}(0)$---the intercept---requires a single measurement on the final system.
This enables prediction of the complete VRAM--latency trade-off curve without exhaustive experimentation.

On the RTX~5070\,Ti, the theoretical slope from profiling is $-229.5$\,ms/layer; the measured slope from linear regression over 29 configurations is $-228.0$\,ms/layer, a difference of 0.7\%.
The residual gap is attributable to implementation-level overhead (Python hook dispatch, CUDA event synchronization) that is eliminated when a layer becomes resident, and is not a fundamental limitation of the prediction model.


\textbf{Validation.} Table~\ref{tab:prediction_validation} compares the predicted and measured inference times across the full resident sweep range.
The prediction error remains within 1.3\% across all configurations, confirming that a single profiling run and one measurement suffice to predict the entire VRAM--latency trade-off curve.

\begin{table}[t]
\centering
\caption{Predicted vs.\ measured inference time for varying numbers of GPU-resident VLM layers (RTX~5070\,Ti).}
\label{tab:prediction_validation}
\begin{tabular}{rrrr}
\toprule
\textbf{Resident} & \textbf{Predicted (s)} & \textbf{Measured (s)} & \textbf{Error} \\
\midrule
0  & 10.482 & 10.482 & $\pm 0.00\%$ \\
5  &  9.335 &  9.360 & $-0.27\%$ \\
10 &  8.187 &  8.219 & $-0.39\%$ \\
15 &  7.040 &  7.090 & $-0.72\%$ \\
20 &  5.892 &  5.932 & $-0.66\%$ \\
25 &  4.745 &  4.803 & $-1.22\%$ \\
28 &  4.056 &  4.092 & $-0.87\%$ \\
\bottomrule
\end{tabular}
\end{table}

\section{Evaluation}
\label{sec:evaluation}

\subsection{Experimental Setup}

All experiments use NVIDIA's Alpamayo-R1-10B (21.52\,GB, BF16) with 10 diffusion steps.
Each configuration is measured over 30 trials with one warmup run; we report the mean after discarding the first trial.
GPU clocks are locked at their maximum frequency via \texttt{nvidia-smi -lgc} to eliminate dynamic frequency scaling.
All host-side parameter tensors use pinned memory (\texttt{pin\_memory()}) for optimal DMA throughput.

\textbf{Hardware.} The primary evaluation platform is the RTX~5070\,Ti (16\,GB VRAM, PCIe Gen5~x16, DDR5-5600 single-channel) running Ubuntu with Linux kernel~6.8.
Cross-platform validation uses the RTX~3080\,Ti (12\,GB VRAM, PCIe Gen3~x16).

\textbf{Baselines.} We compare against two baselines: (1)~Hugging Face Accelerate's \texttt{device\_map="auto"}~\cite{accelerate2022} with a 15\,GB VRAM limit (14.52\,s, 14.45\,GB peak), representing the standard practitioner approach, and (2)~the preloading baseline estimated as $C^{\text{opt}} = 2.86$\,s on the RTX~5070\,Ti and $C^{\text{opt}} = 5.86$\,s on the RTX~3080\,Ti.

\begin{figure}[t]
\centering
\includegraphics[width=\columnwidth]{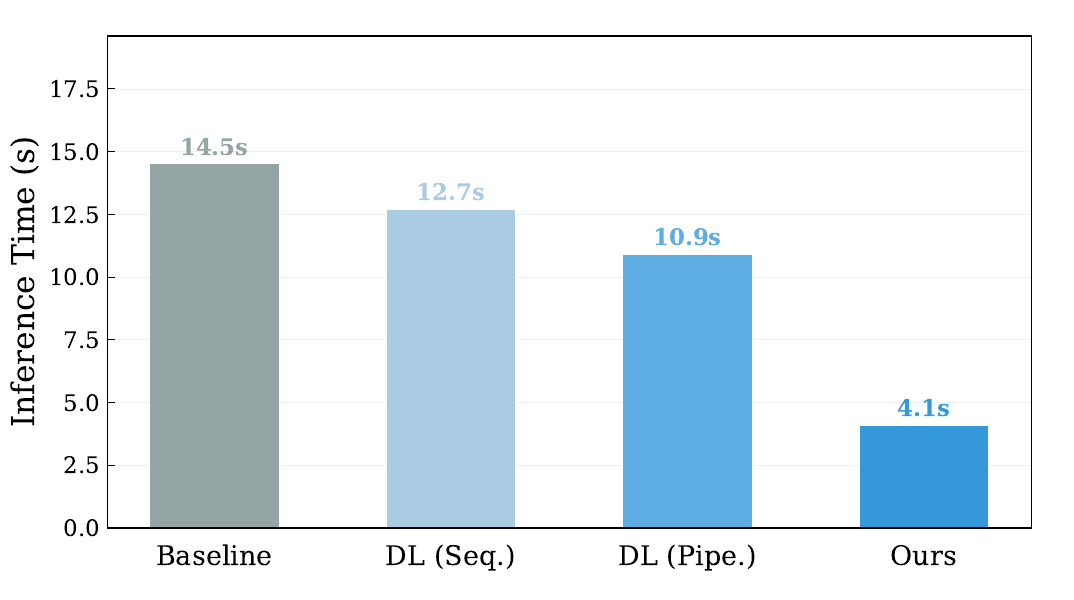}
\caption{Progressive optimization: Baseline $\to$ Sequential DL $\to$ Pipelined DL $\to$ Ours ($N_r{=}28$). Each stage improves latency; resident layer selection provides the largest gain (2.67$\times$).}
\label{fig:eval_progression}
\end{figure}

\subsection{Progressive Optimization}

Fig.~\ref{fig:eval_progression} shows the inference time reduction across the three optimization stages of our method.
Sequential Demand Layering reduces inference time from 14.52\,s to 12.72\,s (1.14$\times$) while cutting VRAM from 14.45\,GB to 3.99\,GB---a 72\% reduction.
Pipelined Demand Layering with a Double Flat Buffer further reduces time to 10.93\,s (1.33$\times$) by hiding DMA transfers within EXE-intensive module execution.
Finally, with 28 GPU-resident VLM layers ($N_r{=}28$), our method achieves 4.09\,s (3.55$\times$), approaching the theoretical lower bound of 2.86\,s.

\begin{figure}[t]
\centering
\includegraphics[width=\columnwidth]{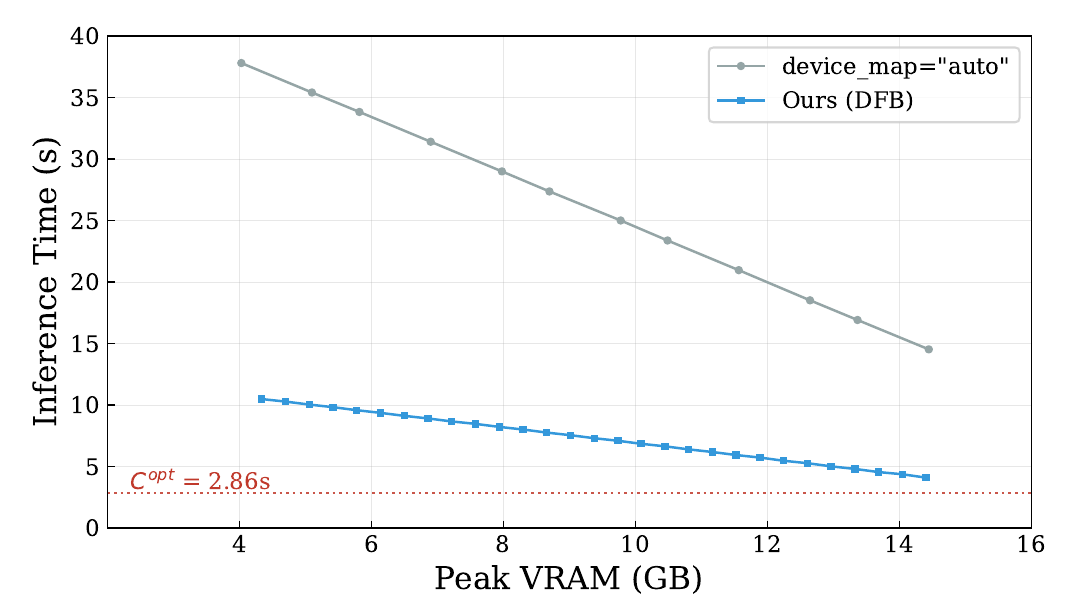}
\caption{VRAM--latency tradeoff. Our method consistently achieves 3.5--3.6$\times$ lower latency than Accelerate offloading at every VRAM budget.}
\label{fig:eval_main}
\end{figure}

\subsection{VRAM--Latency Tradeoff}

\begin{figure*}[t]
      \centering
      \begin{subfigure}[t]{0.24\textwidth}
          \centering
          \includegraphics[width=\linewidth]{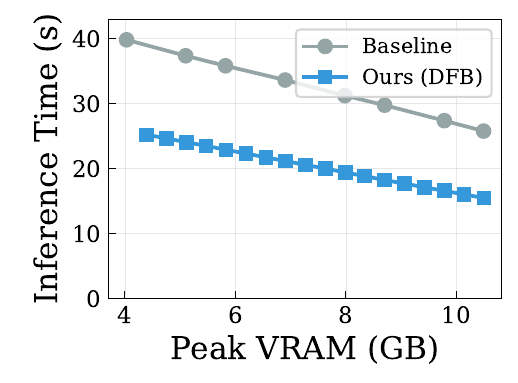}
          \caption{Alpamayo trade-off.}
          \label{fig:alpamayo_tradeoff_3080ti}
      \end{subfigure}
      \hfill
      \begin{subfigure}[t]{0.24\textwidth}
          \centering
          \includegraphics[width=\linewidth]{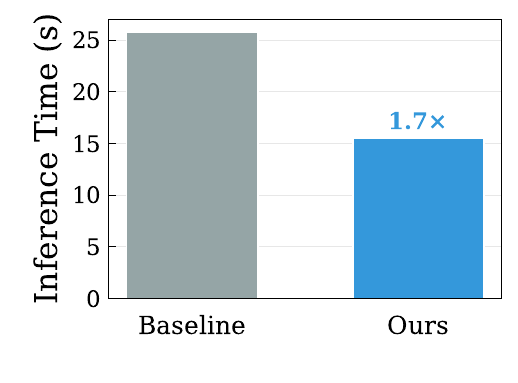}
          \caption{Alpamayo best configuration.}
          \label{fig:alpamayo_best_3080ti}
      \end{subfigure}
      \hfill
      \begin{subfigure}[t]{0.24\textwidth}
          \centering
          \includegraphics[width=\linewidth]{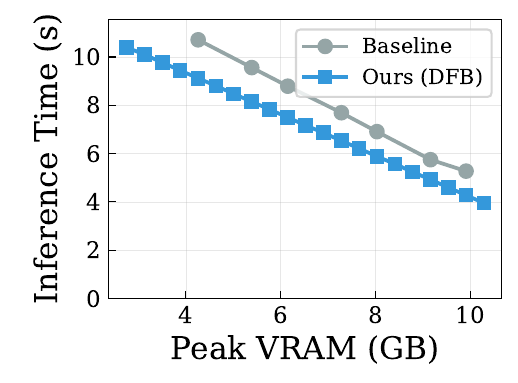}
          \caption{OpenVLA trade-off.}
          \label{fig:openvla_tradeoff_3080ti}
      \end{subfigure}
      \hfill
      \begin{subfigure}[t]{0.24\textwidth}
          \centering
          \includegraphics[width=\linewidth]{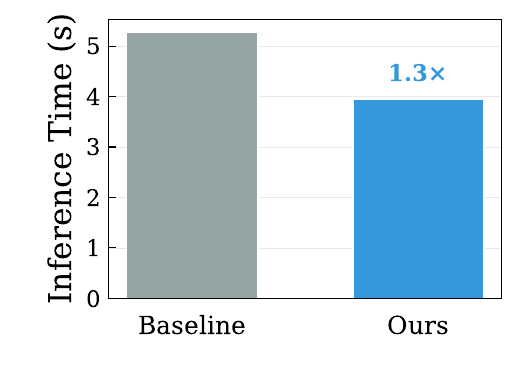}
          \caption{OpenVLA best configuration.}
          \label{fig:openvla_best_3080ti}
      \end{subfigure}
      \caption{Comparison of baseline and DFB on RTX 3080 Ti.}
      \label{fig:3080ti_comparison}
\end{figure*}

Fig.~\ref{fig:eval_main} compares the VRAM--latency tradeoff of our method against Accelerate offloading across the full operating range.
At every VRAM budget, our method achieves 3.5--3.6$\times$ lower latency than the baseline.
Notably, Accelerate offloading requires 35.41\,s at 5\,GB VRAM, while our method achieves 10.48\,s at 4.3\,GB, corresponding to 3.4$\times$ lower latency.
At the maximum configuration ($N_r{=}28$, 14.41\,GB), our method reaches 4.09\,s, which is 3.55$\times$ faster than the baseline at comparable VRAM.

\subsection{Numerical Correctness}

Table~\ref{tab:accuracy} verifies that Demand Layering with Double Flat Buffer produces bit-exact output regardless of the number of resident layers.
We compare the predicted trajectory (64 waypoints $\times$ 3 coordinates) against the baseline (full GPU preloading) on the RTX~3090.
The maximum absolute error is zero across all configurations, confirming that our optimization introduces no numerical degradation.

\begin{table}[t]
\centering
\caption{Numerical correctness: max absolute trajectory error between baseline (full GPU) and DFB at varying $N_r$ (RTX~3090).}
\label{tab:accuracy}
\begin{tabular}{rcc}
\toprule
$N_r$ & \textbf{Max Abs.\ Error} & \textbf{Bit-exact} \\
\midrule
0  & 0.0 & \checkmark \\
5  & 0.0 & \checkmark \\
10 & 0.0 & \checkmark \\
15 & 0.0 & \checkmark \\
20 & 0.0 & \checkmark \\
\bottomrule
\end{tabular}
\end{table}

\subsection{Cross-Platform and Cross-Model Scalability}

To verify that our analysis framework generalizes beyond a single GPU and model, we evaluate on an RTX~3080\,Ti (12\,GB VRAM, PCIe Gen3~x16) with both Alpamayo-R1-10B and OpenVLA-7B.


\textbf{Cross-Platform Profiling.} Table~\ref{tab:cross_platform} compares the per-module DMA/EXE ratio $r$ between the RTX~5070\,Ti (PCIe Gen5) and RTX~3080\,Ti (PCIe Gen3).

\begin{table}[t]
\centering
\caption{Cross-platform DMA/EXE ratio comparison. Lower PCIe bandwidth shifts more modules toward DMA-intensive.}
\label{tab:cross_platform}
\begin{tabular}{lcccc}
\toprule
\textbf{Module} & \multicolumn{2}{c}{\textbf{RTX 5070\,Ti (Gen5)}} & \multicolumn{2}{c}{\textbf{RTX 3080\,Ti (Gen3)}} \\
 & $C_{\text{DMA}}$ & $r$ & $C_{\text{DMA}}$ & $r$ \\
\midrule
ViT         & 0.9\,ms  & 0.1  & 2.5\,ms  & 0.24 \\
VLM Prefill & 10.9\,ms & 0.7  & 30.8\,ms & 1.5  \\
VLM Decode  & 10.9\,ms & 12.1 & 30.8\,ms & 20.5 \\
Expert      & 3.6\,ms  & 3.5  & 10.1\,ms & 6.7  \\
\bottomrule
\end{tabular}
\end{table}

The EXE-intensive vs.\ DMA-intensive classification is consistent across both platforms: ViT remains EXE-intensive ($r < 1$), while VLM Decode and Expert are DMA-intensive ($r \gg 1$) on both GPUs.
Notably, on the RTX~3080\,Ti with PCIe Gen3, VLM Prefill transitions from EXE-intensive ($r{=}0.7$) to DMA-intensive ($r{=}1.5$), demonstrating that lower PCIe bandwidth shifts more modules toward DMA-intensive behavior, making GPU-resident layer selection even more critical on older platforms.

Fig.~\ref{fig:3080ti_comparison}(a)--(b) show the Alpamayo results on the RTX~3080\,Ti.
At comparable VRAM (${\sim}10.5$\,GB), our method achieves 15.46\,s ($N_r{=}17$) versus the Accelerate baseline's 25.74\,s---a 1.7$\times$ speedup, confirming that the DFB framework transfers effectively to VRAM-constrained PCIe Gen3 platforms.


\textbf{Cross-Model Validation: OpenVLA-7B.} To verify model generality, we apply our framework to OpenVLA-7B~\cite{openvla2024}, a 7B-parameter VLA model with 32 LLM layers.
Fig.~\ref{fig:3080ti_comparison}(c)--(d) show the results on the RTX~3080\,Ti.
At comparable VRAM (${\sim}10$\,GB), our method achieves 3.94\,s ($N_r{=}20$) versus the Accelerate baseline's 5.27\,s---a 1.3$\times$ speedup, demonstrating that the DFB framework and residency policy generalize across different VLA architectures.

The lower speedup on the RTX~3080\,Ti compared to the RTX~5070\,Ti (1.7$\times$ vs.\ 3.55$\times$ for Alpamayo) is expected: PCIe Gen3 bandwidth is roughly 3$\times$ lower than Gen5, increasing $C_\text{DMA}$ and the DMA-intensive bottleneck that our method targets.
Since the baseline also suffers from higher transfer overhead on Gen3, the relative gap narrows, but our method still consistently outperforms the baseline across all VRAM budgets on both platforms.

\section{Discussion}
\label{sec:discussion}

\subsection{Platform-Specific Issues}

During experimentation, we discovered that Linux kernel~6.17 on the Arrow Lake platform causes CPU memory bandwidth to collapse to 3.6\% of theoretical throughput, inflating inference time to 59.09\,s.
Downgrading to kernel~6.8 resolved the issue (inference time reduced to 8.95\,s, a 6.6$\times$ improvement), underscoring that system-level optimization is sensitive to OS and driver layers.
All experiments specify the exact kernel version for reproducibility.

\section{Related Work}
\label{sec:related}

Our work lies at the intersection of GPU memory management, model offloading, and real-time DNN inference optimization.
We organize related work into three categories.

\subsection{Layer-wise Offloading and Memory Virtualization}

vDNN~\cite{vdnn2016} pioneered GPU memory virtualization by dynamically swapping activation data between GPU and CPU during training, achieving up to 95\% memory reduction.
While it established the prefetch-overlap paradigm that our work also employs, vDNN targets training activations rather than inference weights, differing in both data type and execution pattern.

Demand Layering~\cite{demandlayering2022} is the most directly related prior work.
It exploits the layer-by-layer execution pattern of DNNs to load parameters from SSD to GPU one layer at a time, achieving 96.5\% memory reduction with 14.8\% latency overhead through pipelined I/O hiding.
However, the CNNs targeted by Demand Layering are predominantly EXE-intensive ($C_{\text{DMA}} < C_{\text{EXE}}$), allowing pipelining to hide most of the additional transfer delays.
In contrast, VLA models such as Alpamayo contain DMA-intensive modules ($C_{\text{DMA}} \gg C_{\text{EXE}}$, $r \approx 12$ for VLM Decode), where pipelining alone cannot hide the transfer overhead.
Our work analyzes these heterogeneous module characteristics and introduces a residency policy to address the DMA-intensive bottleneck.

RT-Swap~\cite{rtswap2024} provides transparent GPU memory swap scheduling for multi-DNN real-time inference, making 72\% more task sets schedulable when memory demand exceeds capacity by 96.2\%.
While RT-Swap manages memory at the task level for concurrent DNNs, our work manages memory at the layer level within a single large VLA model.

\subsection{Large Model Inference Offloading}

DeepSpeed Inference~\cite{deepspeed} supports multi-GPU Transformer inference through tensor parallelism and model-parallel inference.
Its ZeRO-Inference component provides layer-wise CPU/NVMe offloading with prefetch, but optimizes for throughput with large batch sizes rather than single-request latency for real-time control.

FlexGen~\cite{flexgen2023} aggregates GPU, CPU, and disk memory to run LLMs on a single GPU, using linear programming to optimize tensor placement and achieving 100$\times$ throughput improvement for OPT-175B.
However, FlexGen targets batch throughput optimization, which diverges from the single-inference latency minimization required for real-time VLA deployment.

NEO~\cite{neo2024} offloads attention computation and KV cache from GPU to CPU via asymmetric pipelining, achieving up to 7.5$\times$ throughput improvement on T4 GPUs.
Unlike our approach which offloads layer weights, NEO offloads computation itself and targets uniform Transformer stacks, whereas our VLA model comprises a heterogeneous multi-stage pipeline (vision encoder, language model, and diffusion model).

PowerInfer~\cite{powerinfer2023} exploits neuron activation sparsity in LLMs to enable GPU-CPU hybrid inference, selectively loading only ``hot'' neurons onto the GPU.
While highly effective for sparse-activation LLMs, VLA models like Alpamayo activate all parameters during inference (dense execution). Therefore direct adoption is non-trivial.

llama.cpp~\cite{llamacpp} is a widely deployed CPU/GPU offloading framework for LLM inference on consumer hardware.
It supports partial GPU offloading at layer granularity but does not explicitly provide the prefetching/residency mechanisms targeted in this work.

Hugging Face Accelerate~\cite{accelerate2022} provides an automatic device-mapping option (\texttt{device\_map="auto"}) for static parameter partitioning across GPU/CPU at load time.
However, it does not explicitly expose the prefetching and transfer--compute overlap mechanisms used in our method, resulting in significant performance degradation under VRAM constraints as demonstrated in our analysis (Section~\ref{sec:problem}).

\subsection{Model Compression}

GPTQ~\cite{gptq2023} and AWQ~\cite{awq2024} compress LLM weights to typically 3--4 bits via post-training quantization with minimal accuracy loss.
These approaches trade model precision for memory savings---an orthogonal direction to our system-level optimization that preserves full BF16 precision.
While quantization and our techniques are combinable in principle, this work demonstrates the performance achievable through system optimization alone.

\section{Conclusion}
\label{sec:conclusion}

We presented a framework for memory-efficient VLA model inference on VRAM-constrained commodity GPUs without model modification.
Applied to the 21.52\,GB Alpamayo-R1-10B model on an RTX~5070\,Ti (16\,GB), the framework demonstrates that system-level optimization alone can bridge the memory gap for practical VLA deployment.

Our key contributions are fourfold.
First, we characterized the heterogeneous module behavior of VLA models, classifying layers as EXE-intensive or DMA-intensive and revealing that VLM Decode layers ($r \approx 12$) make transfer hiding via pipelining structurally impossible.
Second, we derived closed-form residency benefit expressions for EXE-intensive, DMA-intensive, and hybrid modules, and proposed an interleaved placement policy that reduces the $2^{36}$ configuration search space to a single profiling run.
Third, we presented a performance prediction model that estimates inference time for arbitrary residency configurations within 1.3\% error from profiling data alone.
Fourth, we demonstrated 3.55$\times$ speedup over Accelerate offloading on the RTX~5070\,Ti with full BF16 precision and validated the analysis framework across multiple GPU platforms.

We acknowledge that the achieved inference time of 4.09\,s on the RTX~5070\,Ti remains far from the sub-100\,ms regime required for closed-loop autonomous driving control.
However, this limitation is fundamental to current VLA model scale rather than to our optimization framework: even on the high-end RTX~5090 (32\,GB) with full GPU preloading and no memory constraints, Alpamayo-R1-10B inference still requires 1.03\,s---an order of magnitude above real-time requirements.
Bridging this gap requires advances in both model architecture (smaller, faster VLA models) and hardware (higher compute throughput), which is the fundamental challenge for VLA-based autonomous driving.
Our contribution lies in ensuring that, for any given model and GPU, inference latency approaches the theoretical optimum $C^{\text{opt}}$ as closely as possible under VRAM constraints.

\balance
\bibliographystyle{IEEEtran}
\bibliography{refs}

@article{alpamayo2025,
    title={Alpamayo-R1: Bridging Reasoning and Action Prediction for Generalizable Autonomous Driving in the Long
    Tail},
    author = {Wang, Yan and Luo, Wenjie and Bai, Junjie and others},
    journal = {arXiv preprint arXiv:2511.00088},
    year={2025}
}

@article{rt2_2023,
  author    = {Anthony Brohan and Noah Brown and Justice Carbajal and others},
  title     = {{RT-2}: Vision-Language-Action Models Transfer Web Knowledge to Robotic Control},
  journal   = {arXiv preprint arXiv:2307.15818},
  year      = {2023}
}

@article{drivevlm2024,
  author    = {Xiaoyu Tian and Junru Gu and Bailin Li and Yicheng Liu and Yang Wang and Zhiyong Zhao and Kun Zhan and Peng Jia and Xianpeng Lang and Hang Zhao},
  title     = {{DriveVLM}: The Convergence of Autonomous Driving and Large Vision-Language Models},
  journal   = {arXiv preprint arXiv:2402.12289},
  year      = {2024}
}

@article{openvla2024,
  author    = {Moo Jin Kim and Karl Pertsch and Siddharth Karamcheti and Ted Xiao and Ashwin Balakrishna and Suraj Nair and Rafael Rafailov and Ethan Foster and Grace Lam and Pannag Sanketi and Quan Vuong and Thomas Kollar and Benjamin Burchfiel and Russ Tedrake and Dorsa Sadigh and Sergey Levine and Percy Liang and Chelsea Finn},
  title     = {{OpenVLA}: An Open-Source Vision-Language-Action Model},
  journal   = {arXiv preprint arXiv:2406.09246},
  year      = {2024}
}

@inproceedings{demandlayering2022,
  author    = {Mingoo Ji and Saehanseul Yi and Changjin Koo and Sol Ahn and Dongjoo Seo and Nikil Dutt and Jong-Chan Kim},
  title     = {Demand Layering for Real-Time {DNN} Inference with Minimized Memory Usage},
  booktitle = {Proceedings of the 43rd IEEE Real-Time Systems Symposium (RTSS)},
  year      = {2022},
  pages     = {277--290},
  doi       = {10.1109/RTSS55097.2022.00035}
}

@inproceedings{rtswap2024,
  author    = {Woosung Kang and Jinkyu Lee and Youngmoon Lee and Sangeun Oh and Kilho Lee and Hoon Sung Chwa},
  title     = {{RT-Swap}: Addressing GPU Memory Bottlenecks for Real-Time Multi-DNN Inference},
  booktitle = {Proceedings of the 2024 IEEE 30th Real-Time and Embedded Technology and Applications Symposium (RTAS)},
  year      = {2024},
  pages     = {373--385}
}

@inproceedings{flexgen2023,
  author    = {Ying Sheng and Lianmin Zheng and Binhang Yuan and Zhuohan Li and Max Ryabinin and Beidi Chen and Percy Liang and Christopher Re and Ion Stoica and Ce Zhang},
  title     = {FlexGen: High-Throughput Generative Inference of Large Language Models with a Single GPU},
  booktitle = {Proceedings of the 40th International Conference on Machine Learning},
  year      = {2023},
  pages     = {31094--31116}
}

@misc{neo2024,
  author       = {Xuanlin Jiang and Yang Zhou and Shiyi Cao and Ion Stoica and Minlan Yu},
  title        = {NEO: Saving GPU Memory Crisis with CPU Offloading for Online LLM Inference},
  year         = {2024},
  eprint       = {2411.01142},
  archivePrefix= {arXiv},
  primaryClass = {cs.DC}
}

@misc{accelerate2022,
  author       = {{Hugging Face}},
  title        = {Accelerate},
  year         = {2026},
  howpublished = {\url{https://github.com/huggingface/accelerate}},
  note         = {GitHub repository, accessed: 2026-03-26}
}

@inproceedings{vdnn2016,
  author    = {Minsoo Rhu and Natalia Gimelshein and Jason Clemons and Arslan Zulfiqar and Stephen W. Keckler},
  title     = {{vDNN}: Virtualized Deep Neural Networks for Scalable, Memory-Efficient Neural Network Design},
  booktitle = {Proceedings of the 49th Annual IEEE/ACM International Symposium on Microarchitecture (MICRO)},
  year      = {2016},
  doi       = {10.1109/MICRO.2016.7783721}
}

@misc{cuda_programming_guide,
  author       = {{NVIDIA}},
  title        = {CUDA C++ Programming Guide},
  year         = {2026},
  howpublished = {\url{https://docs.nvidia.com/cuda/cuda-c-programming-guide/}},
  note         = {CUDA Toolkit Documentation, accessed: 2026-03-26}
}

@inproceedings{gptq2023,
  author    = {Elias Frantar and Saleh Ashkboos and Torsten Hoefler and Dan Alistarh},
  title     = {{GPTQ}: Accurate Post-Training Quantization for Generative Pre-Trained Transformers},
  booktitle = {Proceedings of the International Conference on Learning Representations (ICLR)},
  year      = {2023}
}

@inproceedings{awq2024,
  author    = {Ji Lin and Jiaming Tang and Haotian Tang and Shang Yang and Wei-Ming Chen and Wei-Chen Wang and Guangxuan Xiao and Xingyu Dang and Chuang Gan and Song Han},
  title     = {{AWQ}: Activation-aware Weight Quantization for On-Device {LLM} Compression and Acceleration},
  booktitle = {Proceedings of Machine Learning and Systems (MLSys)},
  year      = {2024}
}

@article{survey_e2e_ad2024,
  author    = {Li Chen and Penghao Wu and Kashyap Chitta and Bernhard Jaeger and Andreas Geiger and Hongyang Li},
  title     = {End-to-End Autonomous Driving: Challenges and Frontiers},
  journal   = {IEEE Transactions on Pattern Analysis and Machine Intelligence},
  year      = {2024},
  volume    = {46},
  number    = {12},
  pages     = {10164--10183},
  doi       = {10.1109/TPAMI.2024.3435937}
}

@inproceedings{deepspeed,
  author    = {R. Y. Aminabadi and S. Rajbhandari and M. Zhang and A. A. Awan and C. Li and D. Li and E. Zheng and J. Rasley and S. Smith and O. Ruwase and Y. He},
  title     = {{DeepSpeed-Inference}: Enabling Efficient Inference of Transformer Models at Unprecedented Scale},
  booktitle = {SC22: International Conference for High Performance Computing, Networking, Storage and Analysis},
  year      = {2022},
  pages     = {1--15}
}

@misc{powerinfer2023,
  author       = {Yixin Song and Zeyu Mi and Haotong Xie and Haibo Chen},
  title        = {PowerInfer: Fast Large Language Model Serving with a Consumer-Grade GPU},
  year         = {2023},
  eprint       = {2312.12456},
  archivePrefix= {arXiv}
}

@misc{llamacpp,
  author       = {G. Gerganov},
  title        = {{llama.cpp}},
  year         = {2026},
  howpublished = {\url{https://github.com/ggml-org/llama.cpp}},
  note         = {GitHub repository, accessed: 2026-03-26}
}

\end{document}